\documentclass[conference]{IEEEtran}
\ifCLASSINFOpdf
\else
\fi
\usepackage[utf8]{inputenc}
\usepackage{amsmath}
\usepackage{amsthm}
\usepackage{cite}
\usepackage{mathtools}
\usepackage{amsfonts}
\usepackage{amssymb}
\usepackage{graphicx}
\usepackage{subfig}
\usepackage{float}
\usepackage{graphicx}
\usepackage{mathrsfs}
\usepackage{algorithm}
\usepackage[noend]{algpseudocode}
\usepackage{graphicx}
\usepackage{xcolor}
\usepackage{colortbl}
\usepackage{etoolbox}

\usepackage{pbox}
\usepackage{lipsum}

\DeclareMathOperator*{\argmin}{arg\,min}
\allowdisplaybreaks
\usepackage{algorithm}
\usepackage{algpseudocode}
\usepackage{algpseudocode}
\algnewcommand{\LineComment}[1]{\State \(\triangleright\) #1}
\usepackage[T1]{fontenc}
\usepackage[utf8]{inputenc}
\newcommand{\mathbbm}[1]{\text{\usefont{U}{bbm}{m}{n}#1}}
\IEEEoverridecommandlockouts

\usepackage{verbatim}

\newtheorem{theorem}{\bf Theorem}

\usepackage{geometry}
\def\href#1#2{#1 #2}

\ifCLASSINFOpdf

\else

\fi

\begin{document}
	\setlength{\abovedisplayskip}{0pt}
	\setlength{\belowdisplayskip}{0pt}
	\setlength{\abovedisplayshortskip}{0pt}
	\setlength{\belowdisplayshortskip}{0pt}

\title{Federated Learning in the Sky: Joint Power Allocation and Scheduling with UAV Swarms}

\author{\IEEEauthorblockN{Tengchan Zeng, Omid Semiari, Mohammad Mozaffari, Mingzhe Chen, Walid Saad, and Mehdi Bennis\vspace{-2cm}}
%
%
%
    \thanks{ This research was supported, in part, by the U.S. National Science Foundation under Grants CNS-1739642 and CNS-1941348, and by the Academy of Finland Project CARMA, by the Academy of Finland Project MISSION, by the Academy of Finland Project SMARTER, as well as by the INFOTECH Project NOOR.
	
	T. Zeng and W. Saad are with Wireless@VT, Department of Electrical and Computer Engineering, Virginia Tech, Blacksburg, VA, 24061 USA (e-mail: tengchan@vt.edu; walids@vt.edu).
	
	O. Semiari is with Department of Electrical and Computer Engineering, University of Colorado Colorado Springs, Colorado Springs, CO, 80918 USA (e-mail: osemiari@uccs.edu).
	
	M. Mozaffari is with Ericsson Research, Santa Clara, CA, 95054 USA (e-mail: mohammad.mozaffari@ericsson.com).
	
	M. Chen is with Department of Electrical Engineering, Princeton University, Princeton, NJ, 08544 USA (e-mail: mingzhec@princeton.edu).
	
	M. Bennis is with the Centre for Wireless Communications, University of Oulu, 90014 Oulu, Finland (e-mail:mehdi.bennis@oulu.fi).
}
}

\makeatletter
\patchcmd{\@maketitle}
{\addvspace{0.5\baselineskip}\egroup}
{\addvspace{0.5\baselineskip}\egroup}
{}
{}
\makeatother

\maketitle

\begin{abstract}
	Unmanned aerial vehicle (UAV) swarms must exploit machine learning (ML) in order to
	execute various tasks ranging from coordinated trajectory planning to cooperative target recognition.   
	However, due to the lack of continuous connections between the UAV swarm and ground base stations (BSs), using centralized ML will be challenging, particularly when dealing with a large volume of data.
	In this paper, a novel framework is proposed to implement distributed federated learning (FL) algorithms within a UAV swarm that consists of a leading UAV and several following UAVs. 
	Each following UAV trains a local FL model based on its collected data and then sends this trained local model to the leading UAV who will aggregate the received models, generate a global FL model, and transmit it to followers over the intra-swarm network.   
	To identify how wireless factors, like fading, transmission delay, and UAV antenna angle deviations resulting from wind and mechanical vibrations, impact the performance of FL, a rigorous convergence analysis for FL is performed. 
	Then, a joint power allocation and scheduling design is proposed to optimize the convergence rate of FL while taking into account the energy consumption during convergence and the delay requirement imposed by the swarm's control system.
	Simulation results validate the effectiveness of the FL convergence analysis and show that the joint design strategy can reduce the number of communication rounds needed for convergence by as much as $35$\% compared with the baseline design.  
\end{abstract}

\IEEEpeerreviewmaketitle
\section{Introduction}
Swarms of unmanned aerial vehicles (UAVs) will play an important role in various services ranging from delivery of goods to monitoring \cite{8660516} and \cite{8533634}.
To deliver those services, UAV swarms will employ machine learning (ML) for executing various tasks such as consensus trajectory planning, target recognition, and localization. 
However, due to the high altitude and mobility of UAVs, continuous connections between UAVs and ground base stations (BSs) cannot be guaranteed.
Hence, using centralized ML approaches to execute learning-related tasks will be challenging, particularly when transmitting a large volume of data over aerial links. 
Instead, a distributed learning approach would be more apropos \cite{Wireless_Network_Intelligence_at_the_Edge}. 
In particular, one can use federated learning (FL) to enable each UAV to perform distributed ML tasks without relying on any centralized BSs \cite{DBLP:journals/corr/KonecnyMRR16}. 
In this case, UAVs do not need to send any raw data to BSs when training learning models. 


In essence, FL allows each UAV in a swarm to train its learning model based on its own collected data, and it can use the intra-swarm network to share FL parameters related to the learned models with other UAVs.   
As the learning process proceeds, UAVs in the swarm can reach a consensus on their collective learning tasks, e.g., trajectory planning or target recognition. 
However, since the updates of the learning models in FL are transmitted over a wireless network, the FL convergence and task consensus for the UAV swarm will inevitably be affected by wireless factors such as transmission delay.
Also, due to the high mobility of UAVs, other factors (like wind and mechanical vibrations) can increase the uncertainty of wireless channels by affecting the UAVs' antenna angles which, in turn, will impact the FL convergence.


A number of recent works have investigated how wireless communication impacts FL \cite{chen2020convergence,zeng2019energy,8851249}. 
For instance, in \cite{chen2020convergence}, the authors solve the joint learning, wireless resource allocation, and user selection problem to minimize the FL convergence time while optimizing the FL performance.
Also, the work in \cite{zeng2019energy} proposes a strategy for bandwidth allocation and device scheduling to improve the energy efficiency for networks implementing FL. 
Moreover, \cite{8851249} studies the impact of different scheduling policies on the performance of FL.
While interesting, none of these works in \cite{chen2020convergence,zeng2019energy,8851249} considers the role of FL in a UAV swarm.
Also, due to the high mobility of UAVs and their limited energy, the analysis in \cite{chen2020convergence,zeng2019energy,8851249} cannot be directly applied for UAV swarms.  

The \emph{main contribution} of this paper is a novel framework for enabling FL within a swarm of wireless-connected UAVs.
In particular, we first conduct a convergence analysis for FL to show how wireless factors within the UAV swarm impact the convergence of FL.
We then determine the \emph{convergence round}, defined as the minimum number of communication rounds needed to achieve FL convergence.  
Using this key insight, we formulate an optimization problem that jointly designs the power allocation and scheduling for the UAV swarm network to reduce the FL convergence round.  
In particular, due to the stringent energy limitations of UAVs, we consider the constraint of the energy consumed by learning, communications, and flying during FL convergence.
We also take into account the delay constraint imposed by the control system to guarantee the stability of the UAV swarm. 
To solve the joint design problem, we use a sample average approximation approach from stochastic programming along with a dual method from convex optimization.
\emph{To the best of our knowledge, this is the first work that implements FL for the UAV swarm, studies the impact of wireless factors on the convergence of FL, and optimizes the FL convergence by jointly designing power allocation and scheduling of the UAV network.}
Simulation results validate the convergence analysis of FL and show that the joint design can reduce the convergence round by as much as $35$\% compared with baselines without the joint design.

The rest of the paper is organized as follows. Section \uppercase\expandafter{\romannumeral2} presents the system model for the UAV swarm.  
Section \uppercase\expandafter{\romannumeral3} analyzes the FL convergence and shows the joint system design. 
Section \uppercase\expandafter{\romannumeral4} provides simulation results, and conclusions are drawn in Section \uppercase\expandafter{\romannumeral5}.

\section{System Model}
\begin{figure}[!t]
	\captionsetup[subfloat]{farskip=9pt,captionskip=1pt}
	\centering 
	\subfloat[Communication and learning models.]{%
		\includegraphics[width=2.7in,height=1.7in]{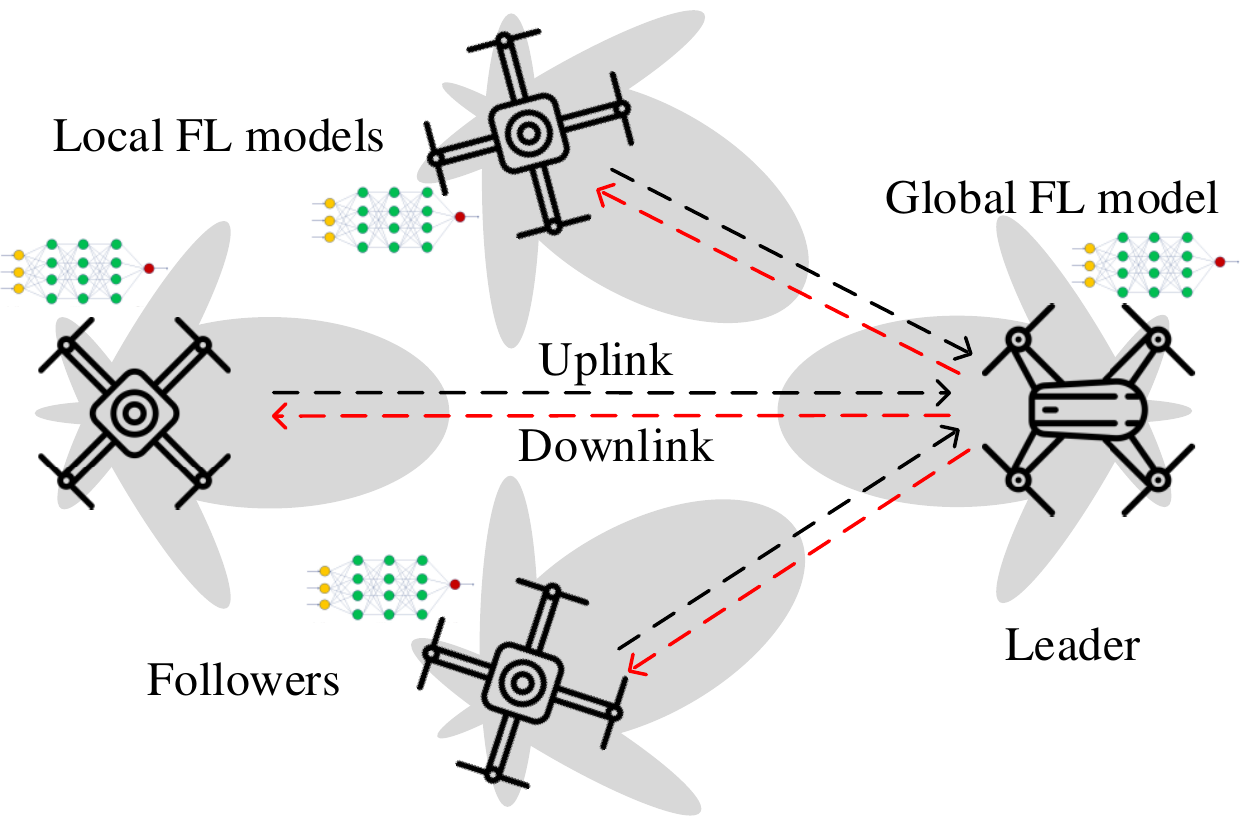}\label{systemmodel1}
		
	}

	\subfloat[Angle deviations and control system.]{%
		\includegraphics[width=2.9in,height=0.8in]{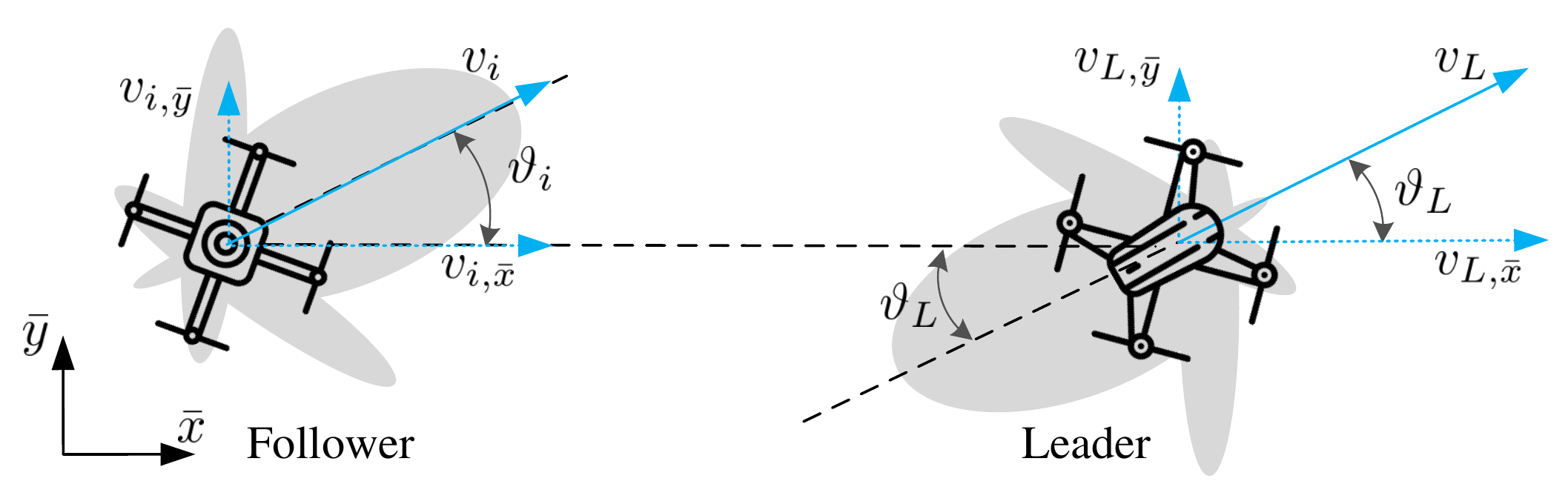}\label{systemmodel2}
	}	
	\caption{\small Illustration of our system model.}
	\label{directionality}
	\vspace{-0.1in}	
\end{figure}

Consider a swarm of wirelessly connected autonomous UAVs flying at the same altitude, as shown in Fig. \ref{directionality}\subref{systemmodel1}. 
The UAV swarm consists of a \emph{leader} $L$ and a set $\mathcal{I}$ of $I$ \emph{followers}. 
Every follower keeps a target distance and speed with the leader. 
While flying, the UAV swarm collects data and performs FL for data analysis and inference tasks like trajectory planning and cooperative target recognition.
Using FL, each follower uses its collected data to train a \emph{local FL model} and send the parameters related to the learned model to the leading UAV in the uplink, as shown in Fig. \ref{directionality}\subref{systemmodel1}.
The leading UAV will integrate all received information to generate a \emph{global FL model}, and, then, transmit the parameters of the global model to following UAVs over the downlink. 
Moreover, to guarantee that the followers fly with the same speed while keeping a safe distance, the leading UAV will also broadcast the target spacing information and its speed and heading direction.


\subsection{Federated learning model}
In the learning model, we assume that UAV $i\!\in\!\mathcal{I}$ collects a set  $\{\boldsymbol{x}_{i1},\boldsymbol{x}_{i2},...,\boldsymbol{x}_{iN_{i}}\}$ of input data where each collected sample is represented by a vector $\boldsymbol{x}_{in}$, $n\!\!\in\!\! \{1,...,N_{i}\}$ that captures the input features and $N_{i}$ is the number of collected samples.
We also assume the input sample $\boldsymbol{x}_{in}$, $n\!\in\! \{1,...,N_{i}\}$, corresponds to a single output $y_{in}$ \cite{DBLP:journals/corr/KonecnyMRR16}.
The output vector is thereby $\{y_{i1},...,y_{iN_{i}}\}$ for UAV $i$. 
We define a vector $\boldsymbol{w}_{i}$ as the parameters related to the local FL model that is trained by $\{\boldsymbol{x}_{i1},\boldsymbol{x}_{i2},...,\boldsymbol{x}_{iN_{i}}\}$ and   $\{y_{i1},...,y_{iN_{i}}\}$ at UAV $i$.
The convergence of the FL training processes requires each local learning vector to converge to a vector $\boldsymbol{w}^{*}$ which solves the following problem:
\begin{align}
\label{FLoptimization_problem}
\argmin_{ \boldsymbol{w}\in \mathbb{R}^{d}} F(\boldsymbol{w}) =  \frac{1}{N}\sum_{i}^{I}\sum_{n=1}^{N_{i}} f(\boldsymbol{w},\boldsymbol{x}_{in},y_{in}),
\end{align}
where $N\!=\! \sum_{i}^{I}N_{i}$ is the total number of the collected samples by all followers, and $f(\boldsymbol{w},\boldsymbol{x}_{in},y_{in})$ captures the loss function when using learning vector $\boldsymbol{w}$ for dataset $\{\boldsymbol{x}_{in},y_{in}\}$.
Note that, the loss function $f(\boldsymbol{w},\boldsymbol{x}_{in},y_{in}), i\! \in\! \mathcal{I}, 0\! \leq\! n \!\leq \!N_{i},$ plays a pivotal role in determining the FL performance, and the expression of the loss function is application-specific. 
For example, for a simple linear regression FL algorithm, $f(\boldsymbol{w},\boldsymbol{x}_{in},y_{in})\!=\!(\boldsymbol{w}^{T}\boldsymbol{x}_{in}\!-\!y_{in})^{2}$.


To solve (\ref{FLoptimization_problem}), the FL framework uses an \emph{iterative update scheme} \cite{DBLP:journals/corr/KonecnyMRR16}.   
In particular, the leading UAV will first generate an initial global FL model represented by vector $\boldsymbol{w}^{(0)}$ and send the initial vector to all followers. 
Hence, in the first communication round, follower $i\!\in\! \mathcal{I}$ will first use $\boldsymbol{w}^{(0)}$ for its own data to train the local model and, then, it sends the vector of the trained model to the leader. 
Next, the leading UAV will aggregate all received local FL vectors and update the global FL model vector which will be later transmitted to the followers.
Each communication round will be followed by another round, and the same process will repeat among leader and followers in each round. 
In this case, as FL proceeds, the local and global models are sequentially updated, and the total loss $F(\boldsymbol{w})$ for the updated global model with vector $\boldsymbol{w}$ will continuously decrease \cite{DBLP:journals/corr/KonecnyMRR16}. 
To identify whether the optimal solution is found for (\ref{FLoptimization_problem}), one must analyze the convergence of the loss function $F(\boldsymbol{w})$ to $F(\boldsymbol{w}^{*})$.
That is, when the gap between the current loss $F(\boldsymbol{w})$ and the minimal loss $F(\boldsymbol{w}^{*})$ is below a threshold $\varepsilon$, the FL optimization problem is solved \cite{8737464}. 
Therefore, we can use the convergence of $F(\boldsymbol{w})$ to $F(\boldsymbol{w}^{*})$ to quantify the FL performance.

Moreover, for each communication round, we can divide the total time duration $T_{r}$ into two periods: Uplink and downlink transmission.
In particular, to guarantee that the leading UAV has enough time to process all received models from its followers, all uplink transmissions should be completed within a target time $T_{u}(\beta)\!=\! \beta T_{r}$, where $\beta \!\in\! \{0,1\}$ is a scheduling parameter to schedule uplink-downlink traffic in time.
Also, to receive the global FL model update from the leading UAV successfully, the time constraint for downlink transmissions is thereby $T_{d}(\beta)\! = \!(1\!-\!\beta) T_{r}$.
In this case, if the communication link between follower $i\in \mathcal{I}$ and leader $L$ fails to meet the time constraints $T_{d}(\beta)$ and $T_{u}(\beta)$, the global FL model cannot use the corresponding  FL model for the aggregation.
At the same time, for the local FL model, the following UAV cannot use the recently updated global vector to train its local data. 
In other words, the transmission delay of the uplink and downlink links will impact the update of the global and local FL models thus having a major impact on FL convergence.

In addition, when training the global FL model, we can calculate the energy consumption for the UAV $L$ as $E_{L}=\kappa C \phi^{2} \sum_{i=1}^{I}S(\boldsymbol{w}_{i})$, where $\kappa$ captures the energy consumption coefficient depending on the computing system and $C$ is the number of computing cycles needed per data bit \cite{8057276}. 
$\phi$ is the frequency of the CPU clock of UAVs, and $S(\boldsymbol{w}_{i})$ is the packet size of $\boldsymbol{w}_{i}$, transmitted from UAV $i\in \mathcal{I}$, in bits.
Similarly, we can determine the training energy consumption for follower $i\in \mathcal{I}$ as $E_{i}=\kappa C \phi^{2} \sum_{n=1}^{N_{i}}S(\boldsymbol{x}_{in})$.


\subsection{Communication model}
To minimize the interference from other UAVs located outside of the swarm, we assume that all UAVs use directional antennas, as shown in Fig. \ref{directionality}\subref{systemmodel1},
However, as shown in Fig. \ref{directionality}\subref{systemmodel2},  due to the impact of wind, payload, and non-ideal mechanical and control systems, the angle of the UAVs will randomly fluctuate and deviate from the initial angle setting. 
Based on the central limit theorem, we model the angle deviation for each UAV as a Gaussian random variable \cite{dabiri2019analytical}. 
Moreover, we consider a squared cosine function to capture the antenna aperture of UAV $j\in \mathcal{I} \cup \{L\}$ when communicating with UAV $l\! \in\!  \mathcal{I} \cup \{L\}/j$ as follows \cite{ITU-R}:
\begin{equation} 
G_{jl}(\theta_{jl}+\vartheta_{j}) = \left\{ 
\begin{array}{c c} \label{aperature}
\cos^{2}(\frac{\pi}{2}(\theta_{jl}+\vartheta_{j})), &\text{if}\hspace{0.1in} |\theta_{jl}+\vartheta_{j}|\leq 1,   \\
G_{\text{min}},  &\text{otherwise,}  \\
\end{array}\right.
\end{equation}
where $\theta_{jl}$ is the initial angle setting for UAV $j$ when communicating with UAV $l$, $\vartheta_{j} \sim \mathcal{N}(0,\sigma_{j}^{2})$ is the angle deviation with variance $\sigma_{j}^{2}$, and $G_{\text{min}}$ captures the antenna gain at the side lobes.
Also, similar to \cite{dabiri2019analytical}, we can approximate (\ref{aperature}) by using a sectionalized expression:
\begin{equation} 
G_{jl}(\theta_{jl}\!+\!\vartheta_{j},M) \!=\! \left\{ 
\begin{array}{c c} 
\cos^{2}(\frac{\pi m}{2 M}),\!\! &\text{if}\hspace{1mm} \frac{m}{M}\!\leq\! |\theta_{jl}\!+\!\vartheta_{j}|\!\leq\! \frac{m+1}{M}, \\
G_{\text{min}},  &\text{otherwise,} \\
\end{array}\right.
\end{equation}
where $m\in\{1,...,M\}$.

To reduce the interference over the uplink transmissions, we assume that uplinks do not share the wireless resource with each other.
Hence, the transmission delay of the uplink between follower $i\!\in\! \mathcal{I}$ and leader $L$ can be calculated as 
\begin{align}
\label{uplink}
T_{iL}= \frac{S(\boldsymbol{w}_{i})}{B_{u} \log_{2} \Big(1+ \frac{p_{i}h_{iL}d_{iL}^{-\alpha}G_{iL}G_{Li}}{\sum_{i' \in \Phi_{i}} p_{i'}h_{i'L}d_{i'L}^{-\alpha}G_{i'L}G_{Li'}+B_{u} \gamma_{0}}   \Big)},
\end{align}
where $B_{u}$ is the bandwidth used by each subchannel in the uplink, $p_{i}\!\in\! (0,p_{\text{max}})$ is the transmission power of UAV $i$ with maximum power as $p_{\text{max}}$, and $\alpha$ is the path-loss exponent. 
$h_{iL}$ is the channel gain of the Rician fading channel between UAVs $i$ and $L$, and $\gamma_{0}$ is the noise power spectral density.  
Note that, despite the use of directional antenna, the swarm still experiences uplink interference generated by UAVs located outside of the swarm. 
In particular, these interfering UAVs share the same channel resource and exist in the main lobe of the UAV $L$, and we define $\Phi_{i}$ as the set of UAVs that generates interference to the uplink from UAV $i$ to UAV $L$.

Similarly, we can derive the transmission delay $T_{Li}$ for the downlink from UAV $L$ to UAV $i\in \mathcal{I}$ 
as: 
\begin{align}
\label{downlink}
T_{Li}= \frac{S(\boldsymbol{w})}{B_{d} \log_{2} \Big(1+ \frac{p_{L}h_{Li}d_{Li}^{-\alpha}G_{Li}G_{iL}}{\sum_{i' \in \Phi_{L}} p_{i'}h_{i'i}d_{i'i}^{-\alpha}G_{i'i}G_{ii'}+B_{d} \gamma_{0}} \Big)},
\end{align}
where $B_{d}$ is the downlink bandwidth, $p_{L}\in (0,p_{\text{max}})$ is the transmission power of UAV $L$, and $\Phi_{L}$ refers to the set of UAVs that will generate interference at the downlink. 

\subsection{Control model}
To guarantee constant speed and altitude and avoid collisions between UAVs within the swarm, the leading UAV will broadcast its speed and heading direction to the followers in the downlink. 
Here, the control system of each follower will use both its sensor data (e.g. location) and information received from the wireless links to coordinate its movement and achieve a target spacing and speed. 
Note that the target distance between the UAV leader and each follower is predefined such that there will be no collision between two nearby UAVs.

Similar to our previous work in \cite{8645472}, we can build a Cartesian coordinate system to capture the locations of UAVs in the swarm, and, then, we decompose the velocity of each UAV into two components, as shown in Fig. \ref{directionality}\subref{systemmodel2}.
We can also define the control law of each UAV the same way as the one provided in \cite{8645472}.  
Since the transmission delay will have a negative impact on the stability control of the UAV swarm, we must consider the delay requirement imposed by the control system when designing the UAV network. 
%

In addition, in order to fly with a constant speed and maintain a stable flying motion, each UAV must spend energy to overcome the gravity and the air drag forces due to the wind and forward motions. 
For a forward speed $v\in (0,v_{\max})$ with $v_{\max}$ as the maximum speed, the minimum flying power of UAV $j \in \mathcal{I} \cup \{L\}$ is  
$\bar{p}_{j,\text{min}}(v) = \hat{v}_{j}  A_{j}$,
where $\hat{v}_{j}$ is the induced velocity required for constant speed $v$ and  given thrust $A_{j}= mg$ with $m$ being the UAV mass and $g$ being the gravitational constant \cite{EnergyConsumption}. 
Also, the induced velocity $\hat{v}_{j}$ can be obtained by solving the following equation \cite{EnergyConsumption}: 
\begin{align}
\hat{v}_{j} = \frac{2 A_{j} }{q r^2 \pi \varrho \sqrt{v^2+ \hat{v}_{j}^2}},
\end{align}
where $q$ and $r$ capture, respectively, the number and diameter of the UAV rotors, and $\varrho$ is the air density. 
Moreover, we can further correct the theoretical minimum motion power consumption by the overall power efficiency $\eta$ of the UAV in order to obtain the actual power consumption as $\bar{p}_{j}(v)\! =\! \bar{p}_{j,\text{min}}(v)/\eta$.
Since the control of a UAV's dynamic motion consumes the most energy \cite{EnergyConsumption}, we must consider the flying energy consumption when designing the swarm of UAVs. 
In particular, the flying energy consumption can be calculated as $\bar{p}_{j}(v) T$ during the flying time $T$.

To guarantee the convergence of FL and the stable operation of the control system in the UAV swarm, we need to properly design the wireless communication network.  
At the same time, to guarantee that the energy spent on learning, communication, and flying will not exceed the energy limitation of each UAV, we need to consider the energy consumption during the FL convergence. 
Next, we first conduct the convergence analysis for the FL algorithm and derive the number of communication rounds needed to achieve the FL convergence. 
Then, we formulate an optimization problem that jointly designs the power allocation and scheduling policy to minimize the convergence round of FL while considering the delay requirement from the control system and energy consumption during the FL convergence.  \vspace{-0.1cm}

\section{Convergence Analysis and Joint Design}
\subsection{FL convergence analysis}
In order to guarantee FL convergence, we assume that the following UAVs adopt a standard gradient descent method to update their local FL models \cite{DBLP:journals/corr/KonecnyMRR16}. 
Thus, for following UAV $i\in \mathcal{I}$, the local model  $\boldsymbol{w}^{(t)}_{i}$ at communication round $t$ is given by
\begin{align}
\label{localFLmodel}
\boldsymbol{w}^{(t)}_{i} = \boldsymbol{w}^{(t-1)} - \frac{\bar{\lambda}}{N_{i}}\triangledown F_{i}(\boldsymbol{w}^{(t-1)}), 
\end{align}
where $\boldsymbol{w}^{(t-1)}$ is the global FL model at communication round $t\!-\!1$, $\bar{\lambda}$ is the learning rate, and $F_{i}(\boldsymbol{w}^{(t-1)}) \!=\! \sum_{n=1}^{N_{i}} f(\boldsymbol{w}^{(t-1)},\boldsymbol{x}_{in},y_{in})$. 
After the leading UAV collects local vectors $\boldsymbol{w}^{(t)}_{i}, i\!\in\! \mathcal{I}$, the global FL model can be updated: 
\begin{align}
\label{globalFLmodel}
\boldsymbol{w}^{(t)} = \frac{\sum_{i=1}^{I}N_{i}\boldsymbol{w}^{(t)}_{i}}{\sum_{i=1}^{I}N_{i}}.
\end{align}
However, for ensuring successful updates of both global and local FL models as shown in (\ref{localFLmodel}) and (\ref{globalFLmodel}), the transmission delay of uplink and downlink should be within, respectively, $T_{u}(\beta)$ and $T_{d}(\beta)$.   
Hence, after considering the impact of the transmission delays, we can rewrite the global FL model update as 
\begin{align}
\label{globalFLmodelupdate}
\boldsymbol{w}^{(t)} = \frac{\sum_{i=1}^{I}N_{i}\boldsymbol{w}^{(t)}_{i}C_{i,t}}{\sum_{i=1}^{I}N_{i}C_{i,t}},
\end{align}
with 
\begin{equation} 
C_{i,t}= \left\{ 
\begin{array}{c c} 
1,&\text{with probability}\hspace{0.1cm} \mathbb{P}(T_{iL,t}\!\leq\! T_{u}(\beta),T_{Li,t}\!\leq\! T_{d}(\beta)), \\
0,&\text{otherwise}. \nonumber \\
\end{array}\right.
\end{equation}
With the aim of quantifying the convergence of FL, we use the notion of a \emph{convergence round}, defined as the minimum number of communication rounds needed to achieve a target difference $\varepsilon$ of the expected gap between current loss and the minimal loss, i.e., $\mathbb{E} (F(\boldsymbol{w})-F(\boldsymbol{w}^{*}))\leq \varepsilon$.
Moreover, to determine the convergence round, we make the following two standard assumptions: Function $F(\boldsymbol{w})$$:\mathbb{R}^{n}\rightarrow \mathbb{R}$ is continuously differentiable, and the gradient of $F(\boldsymbol{w})$ is uniformly Lipschitz continuous with positive parameter $U$.
We also consider the function $F$ to be strongly convex with positive parameter $\mu$,
and these exists constants $\zeta_{1} \geq 0$ and $\zeta_{2}\geq 1$, meeting $||\triangledown F_{i}(\boldsymbol{w})||^2 \leq \zeta_{1} + \zeta_{2} ||\triangledown F(\boldsymbol{w})||^2$ \cite{bertsekas1996neuro}. 
Given the above assumptions, we can derive the convergence round.
\begin{theorem}
	\label{ref1}
	\emph{To realize an expected convergence of $F(\boldsymbol{w})$ under an accuracy threshold $\varepsilon$, i.e., $\mathbb{E} (F(\boldsymbol{w})-F(\boldsymbol{w}^{*}))\leq \varepsilon$, the convergence round is given by:}
	\begin{align}
		\label{something}
		\varphi = \Bigg\lceil \log_{1-\rho} \frac{\varepsilon}{\sum_{i=1}^{I}\sum_{n=1}^{N_{i}}f(\boldsymbol{w}^{(0)},\boldsymbol{x}_{in},y_{in})}  \Bigg\rceil,
		\end{align}
		\emph{where $\lceil \cdot \rceil$ is the ceiling function, and $\rho$ captures the convergence speed given as follows}
	\begin{align}
		\label{rho}
		\rho\! =\!  \frac{\sum_{i=1}^{I}N_{i}\mathbb{P}(T_{iL,t}\!\leq\! T_{u}(\beta),T_{Li,t}\!\leq\! T_{d}(\beta)) \mu}{NU}.
		\end{align}
	\begin{proof}[Proof:\nopunct]Due to the space limitation, the proof is included in Appendix \ref{prooffortheorem6}.
	\end{proof}
\end{theorem}
As shown in Theorem \ref{ref1}, the convergence performance of FL depends on the transmission delay of both uplink and downlink in the network.  
In particular, to increase the convergence speed, we need to maximize the probability that both uplink and downlink meet the corresponding delay requirements of FL.
Thus, Theorem \ref{ref1} provides a concrete characterization of the interplay between wireless communications and FL performance in a UAV swarm.

For the stability analysis of the control system, we will follow the method provided by our previous work in \cite{8645472}.
That is, we first build the augmented error state vector.
Then, we use Lyapunov-Razumikhin theorem to derive the control system delay requirements $\tau_{i}, i \!\in\! \mathcal{I}$, for downlink that can guarantee the stability of the UAV swarm.

\subsection{Problem formulation and solution concept}
Here, we formulate an optimization problem to minimize the convergence round by jointly designing the power allocation and scheduling for the UAV network, as follows:
\begin{align}
\setlength{\abovedisplayskip}{3 pt}
\setlength{\belowdisplayskip}{3 pt}
&\min_{\{\boldsymbol{p},p_{L},\beta,v\}}  \varphi
\label{OptControl}\\
&\hspace{0.02in} \text{s.t.}\hspace{0.05in}  \mathbb{P}\Big[ \varphi E_{L}\!+\!\varphi p_{L}T_{d}(\beta)\!+\!\varphi \bar{p}_{L}(v)T_{r}\!\leq\! \bar{E}  \Big] \!\geq\! \xi_{L}, \label{Con1} \\
&\hspace{0.21in} \mathbb{P}\Big[ \varphi E_{i}\!+\!\varphi p_{i}T_{iL}\!+\!\varphi \bar{p}_{i}(v)T_{r}\!\leq\! \bar{E}  \Big] \!\!\geq \!\xi_{i}, i \!\in \!\mathcal{I}, \label{Con2} \\ 
&\hspace{0.21in} \mathbb{P}(T_{Li}\leq \tau_{i}) \geq \xi_{C}, i \in \mathcal{I}, \label{Con3} \\
&\hspace{0.21in} p_{L} \!\in\! (0,p_{L,\text{max}}), p_{i} \!\in\! (0,p_{i,\text{max}}), i\! \in\! \mathcal{I},
\label{Con4}\\
& \hspace{0.21in}\beta\in (0,1), v \!\in\! (0,v_{\text{max}}),  \label{Con5} 
\end{align}
where vector $\boldsymbol{p}=[p_{1},...,p_{I}]$.
Constraint  (\ref{Con1}) guarantees that the probability of total energy consumption for the leading UAV being less than a threshold $\bar{E}$ will be greater than $\xi_{L} \in (0,1)$. 
Similarly to (\ref{Con1}), constraint (\ref{Con2}) represents the constraint on energy consumption of each follower $i\!\in\! \mathcal{I}$.
Constraint (\ref{Con3}) guarantees that the UAV communication network is reliable to support the stability of the swarm with probability $\xi_{C}$.
Constraints (\ref{Con4}) and (\ref{Con5}) ensure that the optimization variables, i.e., the transmission power, scheduling parameter, and velocity, are chosen within reasonable ranges. 
Note that, in the optimization problem, we also optimize the operation speed of the UAV swarm to minimize the motion energy consumption and relax the energy constraints in (\ref{Con1}) and (\ref{Con2}).

Since both exponent and base in the logarithm function (\ref{something}) are less than $1$, minimizing the logarithm function in (\ref{OptControl}) is equivalent to minimizing the base for the constant exponent.
Also, according to (\ref{rho}), we can simplify (\ref{OptControl}) as
\begin{align}
\label{prob}
\max_{\{\boldsymbol{p},p_{L},\beta,v\}}\sum_{i=1}^{I}N_{i}\mathbb{P}(T_{iL,t}\!\leq\! T_{u}(\beta),T_{Li,t}\!\leq\! T_{d}(\beta)).
\end{align}
We observe that, after simplifications, both objective function and
constraints are represented by probability terms. 
In this case, directly deriving the probability terms will be challenging since it requires multidimensional integrations. 
Also, as the optimization problem is not convex, employing convex approximations to simplify the optimization problem will be impossible. 
Instead, we use a \emph{sample} \emph{average approximation} approach where the probability terms in the objective function and constraints are replaced by an empirical distribution found by random samples \cite{pagnoncelli2009sample}. 
In particular, we first generate $K$ independent samples of the random parameters, i.e., wireless channel gains and angle deviations, and we calculate the corresponding transmission delay and convergence round.
Then, we can reformulate the optimization problem as   
\begin{align}
	&\max_{\{\boldsymbol{p},p_{L},\beta,v\}}  \sum_{i=1}^{I}\sum_{k=1}^{K}N_{i}\mathbbm{1}(T_{u}(\beta)-T_{iL,k})\mathbbm{1}(T_{d}(\beta)-T_{Li,k})
	\label{OptControl1}  \\
	&\text{s.t.}\hspace{-0.03in}  \sum_{k=1}^{K}\mathbbm{1}\big(\bar{E} -    (\varphi_{k} E_{L}\!+\!\varphi_{k} p_{L}T_{d}(\beta)\!+\!\varphi_{k} \bar{p}_{L}(v)T_{r})\big)\geq K\xi_{L}, \label{Con11} \\
	&\hspace{0.13in} \sum_{k=1}^{K}\!\mathbbm{1}\!\big(\bar{E}\! -\! (\varphi_{k} E_{i}\!+\!\varphi_{k} p_{i}T_{iL,k}\!+\!\varphi_{k} \bar{p}_{i}(v)T_{r})  \big)  \!\geq \!K\xi_{i}, i \!\in \!\mathcal{I}, \label{Con21} \\ 
	&\hspace{0.13in} \sum_{k=1}^{K}\mathbbm{1}(\tau_i - T_{Li,k}) \geq K\xi_{C}, i \in \mathcal{I}, \label{Con311} \\
	&\hspace{0.13in} (\ref{Con4})\hspace{0.05in} \text{and}\hspace{0.05in} (\ref{Con5}),  \nonumber 
	\end{align}
	where the indicator function $\mathbbm{1}(r)\!\!=\!\!1$, once $r\!\!\geq\!\! 0$; otherwise, we have $\mathbbm{1}(r)\!=\!0$. 
Due to the presence of the indicator function, the reformulated problem is non-smooth. 
To obtain a smooth problem, we can further replace the indicator functions with modified sigmoid functions, i.e., $\Gamma(r)\!\!= \!\!\frac{1}{1\!+\!\exp(-\bar{c}r)}$, where $\bar{c}$ determines how quickly the modified sigmoid function changes near $0$.
To obtain a sub-optimal solution to the reformulated optimization problem with the indicator functions replaced by the modified sigmoid functions, we can use the \emph{dual} \emph{method} \cite{1658226}.
In particular, the Lagrangian function is
\par \nobreak 
{\small
	\begin{align}
	&\mathcal{J}(\boldsymbol{\lambda},\boldsymbol{p},p_{L},\beta,v) =\sum_{i=1}^{I}\sum_{k=1}^{K}N_{i}\Gamma(T_{u}(\beta)\!-\!T_{iL,k})\Gamma(T_{d}(\beta)\!-\!T_{Li,k})+\nonumber\\
	&\lambda_{1}\Big(\sum_{k=1}^{K}\Gamma\big(\bar{E} -    (\varphi_{k} E_{L}\!+\!\varphi_{k} p_{L}T_{d}(\beta)\!+\!\varphi_{k} \bar{p}_{L}(v)T_{r})\big)-K\xi_{L}\Big)+\nonumber \\ 
	&\sum_{i=1}^{I}\lambda_{i+1}\Big( \sum_{k=1}^{K}\Gamma\big(\bar{E} - (\varphi_{k} E_{i}\!+\!\varphi_{k} p_{i}T_{iL,k}\!+\!\varphi_{k} \bar{p}_{i}(v)T_{r})  \big)-K\xi_{i} \Big) + \nonumber \\
	&\sum_{i=1}^{I}\lambda_{I+1+i}\Big(\sum_{k=1}^{K}\Gamma(\tau_{i} - T_{Li,k})-K\xi_{C}\Big), 
	\end{align}}where vector $\boldsymbol{\lambda}=[\lambda_{1},...,\lambda_{2I+1}] \succeq \boldsymbol{0}_{1 \times(2I+1)}$ is the vector of Lagrangian multipliers, and the dual objective function can be defined as $\mathcal{D}(\boldsymbol{\lambda})=\max_{\boldsymbol{p},p_{L},v,\beta} \mathcal{J}(\boldsymbol{\lambda},\boldsymbol{p},p_{L},\beta,v)$.
The corresponding dual optimization problem is 
	\begin{align}
	\min_{\boldsymbol{\lambda}} \mathcal{D}(\boldsymbol{\lambda})\hspace{0.15in} \text{s.t.}\hspace{0.05in} \boldsymbol{\lambda}\geq \boldsymbol{0}. \label{dualOptimization} 
	\end{align}
	Although the dual problem in (\ref{dualOptimization}) is always convex \cite{nesterov2018lectures}, $\mathcal{D}(\boldsymbol{\lambda})$ is not differentiable. 
Instead, we can use subgradients given by 
\begin{align}
	&\Delta \lambda_{1}\! =\! \sum_{k=1}^{K}\!\Gamma\big(\bar{E} \!- \!   (\varphi^{*}_{k} E_{L}\!+\!\varphi^{*}_{k} p_{L}T^{*}_{d}\!+\!\varphi^{*}_{k} \bar{p}^{*}_{L}T_{r})\big)\!-\!K\xi_{L}, \nonumber \\ 
	&\Delta \lambda_{i+1}\!\! =\!\!  \sum_{k=1}^{K}\!\Gamma\big(\bar{E} \!-\! (\varphi^{*}_{k} E_{i}\!+\!\varphi^{*}_{k} p^{*}_{i}T^{*}_{iL,k}\!+\!\varphi_{k}^{*} \bar{p}^{*}_{i}T_{r})  \big)\!\!-\!\!K\xi_{i}, i\!\in\! \mathcal{I},\nonumber \\ 
	&\Delta \lambda_{I+1+i} =\sum_{k=1}^{K}\Gamma(\tau_i - T_{Li,k}^{*})-K\xi_{C}, i\in \mathcal{I},
	\end{align}
	where the terms $\varphi^{*}_{k}, T^{*}_{d}, T^{*}_{iL,k}, T_{Li,k}^{*}, \bar{p}^{*}_{L}, \bar{p}^{*}_{i}$ are expressed by optimized variables $\boldsymbol{p}^{*},p^{*}_{L},\beta^{*},v^{*}$.
The proof of subgradients is similar to the one provided in \cite{1658226}, and is omitted here.
Thereby, we can solve the problem in (\ref{dualOptimization}) by either the subgradient method or the ellipsoid method, and their complexities are, respectively, $\mathcal{O}\big(\frac{2I+1}{\epsilon^2}\big)$ and $\mathcal{O}\big((2I+1)^2\ln\frac{1}{\epsilon}\big)$ with accuracy $\epsilon$ \cite{nesterov2018lectures}.
Then, the sub-optimal solution of $\{\boldsymbol{p},p_{L},\beta,v\}$ can be obtained by solving dual objective function $\mathcal{D}(\boldsymbol{\lambda})$. 
In particular, similar to \cite{1658226}, we use the iterative method  to sequentially derive the sub-optimal value of each element in $\{\boldsymbol{p},p_{L},\beta,v\}$ (the details are omitted here due to space limitations).
Note that, we assume that all these steps of solving the optimization problem are done by a central unit (e.g., cloud or BS), before the swarm starts training their learning models in FL.
In particular, there is no need for the central unit to collect any information from UAVs, since all samples of wireless channel gains and antenna deviations are randomly generated by the central unit itself.    
Also, since the number of UAVs in the swarm is usually small, the complexity of using sample average approximation and dual approach will be low.
As a result, the central unit can readily obtain the sub-optimal solution to the joint design problem and later send the power allocation and scheduling parameters to UAVs in the swarm.


\begin{table}[!t]
	\normalsize
	\begin{center}
		\caption{ Simulation parameters.}
		\vspace{-0.2cm}
		\label{table_example}
		\resizebox{7.5cm}{!}{
			\begin{tabular}{|c|c|}
				\hline
				\textbf{Parameters} & \textbf{Values} \\ \hline	
				Number of followers $I$ & $5$  \\ \hline
				Transmission power threshold $p_{\max}$     & $0.5$~W    \\ \hline
				Maximum speed $v_{\max}$ & $20$~m/s \cite{8434285} \\ \hline 
				Energy consumption efficient  $\kappa$ &  $10^{-28}$ \cite{8434285} \\ \hline 
				Number of cycles needed per bit $C$ & $10^3$ \cite{8434285} \\ \hline 
				Frequency of the CPU $\phi$ & $10^9$~cycle/s \\ \hline 
				Time for each communication round $T_{r}$ & $0.1$~s \\ \hline 
				Side lobe gain $G_{\min}$, path loss exponent $\alpha$     & $-2$~dB, $2.5$ \\ \hline
				Noise spectral density $\gamma_{0}$ & $-174$~dBm/Hz \\ \hline 
				Packet size $S_{\boldsymbol{w}}$ and $S_{\boldsymbol{w}_{i}}$ & $10$~kB \\ \hline
				Number of rotors $q$ and the diameter $r$ & $4$, $0.254$~m \cite{DBLP:journals/corr/KonecnyMRR16} \\ \hline
				Power efficiency $\eta$ and density $\varrho$ of the air  & $70$~\%, $1.225$~kg/${\text{m}}^3$ \cite{DBLP:journals/corr/KonecnyMRR16}\\ \hline
				Number of samples $K$, 	Energy limits $\bar{E}$ & $1,000$,  $7,000$~J \\ \hline   
		\end{tabular}}
	\end{center}\vspace{-0.6cm}
\end{table}
\section{Simulation Results and Analysis}
For our simulations, we first validate the theoretical analysis in Theorem \ref{ref1}.
Then, we show the impact of angle deviations on the convergence of FL, and we compare our joint design with baseline schemes that optimize power allocation and scheduling separately. 
In particular, we consider two baselines. 
The first baseline is a system with optimized power allocation (same power allocation in the joint design) and randomized scheduling parameters.
The second baseline is a system with optimized scheduling (same scheduling used by the joint design) and randomized power allocation.  
We also assume equal uplink and downlink bandwidths, i.e., $B_{u}\!=\!B_{d}\!=\!1$~MHz, and equal angle deviation variance for each UAV, i.e., $\sigma^{2}_{j}=\sigma^{2}, j \in \mathcal{I}\cup\{L\}$.
All simulation parameters are summarized in Table \ref{table_example}.

\begin{figure}[!t]
	\centering
	\includegraphics[width=2.4in,height=1.8in]{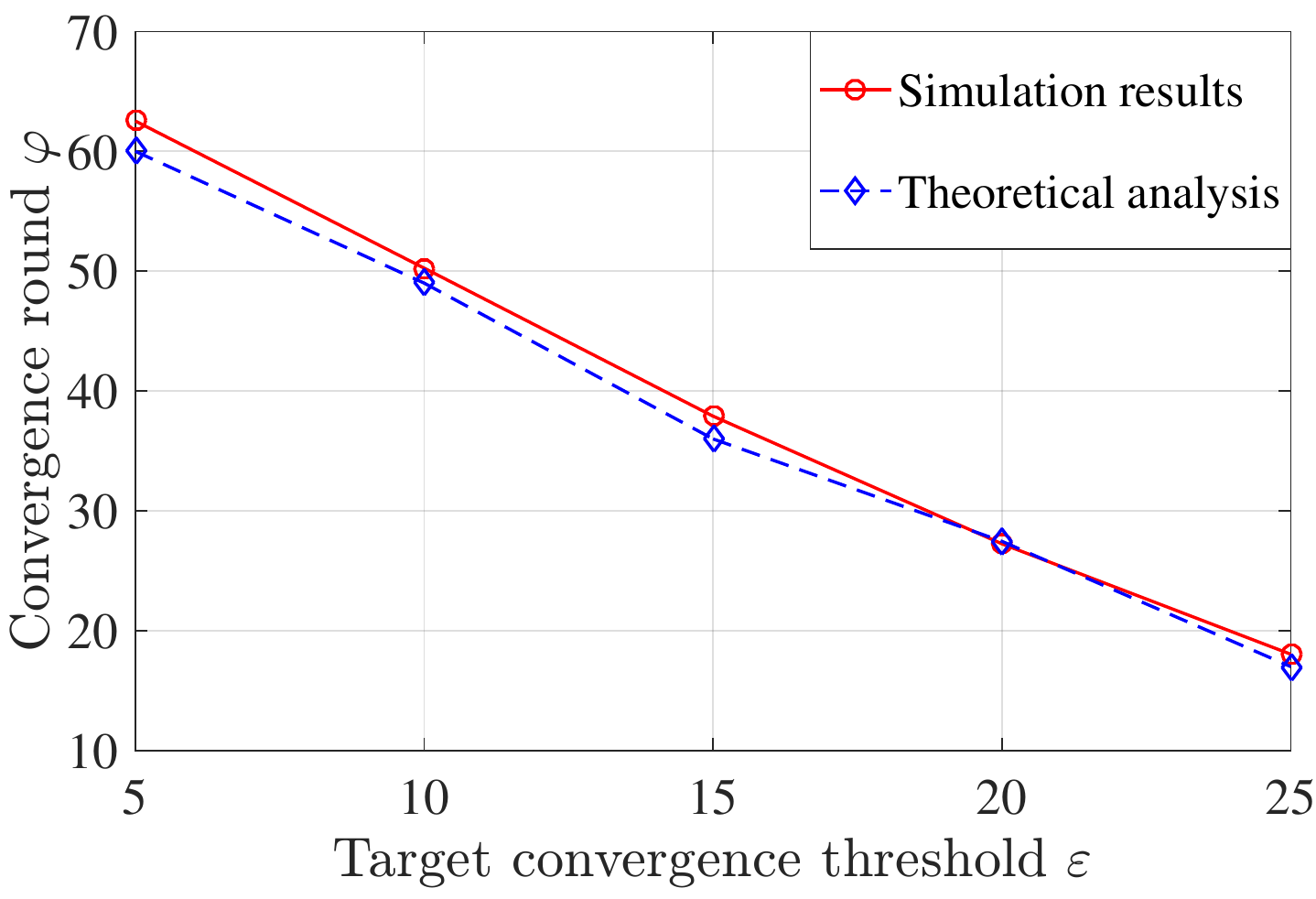}
	\DeclareGraphicsExtensions.
	\caption{\small Validation of Theorem \ref{ref1}.}
	\label{verify}
\end{figure}

Fig. \ref{verify} shows the convergence round versus the difference threshold $\varepsilon$.  
Note that, in Fig. \ref{verify}, we choose the range of $\varepsilon \in (5,25)$ based on the value of $\sum_{i=1}^{I}\sum_{n=1}^{N_{i}}f(\boldsymbol{w}_{0},\boldsymbol{x}_{in},y_{in})$, and the range of $\varepsilon$ will be varied for different settings of data and initial global FL model and the accuracy requirement. 
As observed from Fig. \ref{verify}, the theoretical analysis derived in Theorem \ref{ref1} is aligned with the simulation results with less than $5$\% difference, thus corroborating the validity of Theorem \ref{ref1}.
Moreover, Fig. \ref{verify} shows that, when the difference threshold increases, the convergence round decreases. 
This is because, with a larger difference threshold, the requirement of convergence becomes less stringent. 
In this case, FL requires fewer communication rounds to converge.

\begin{figure}[!t]
	\centering
	\includegraphics[width=2.6in,height=1.75in]{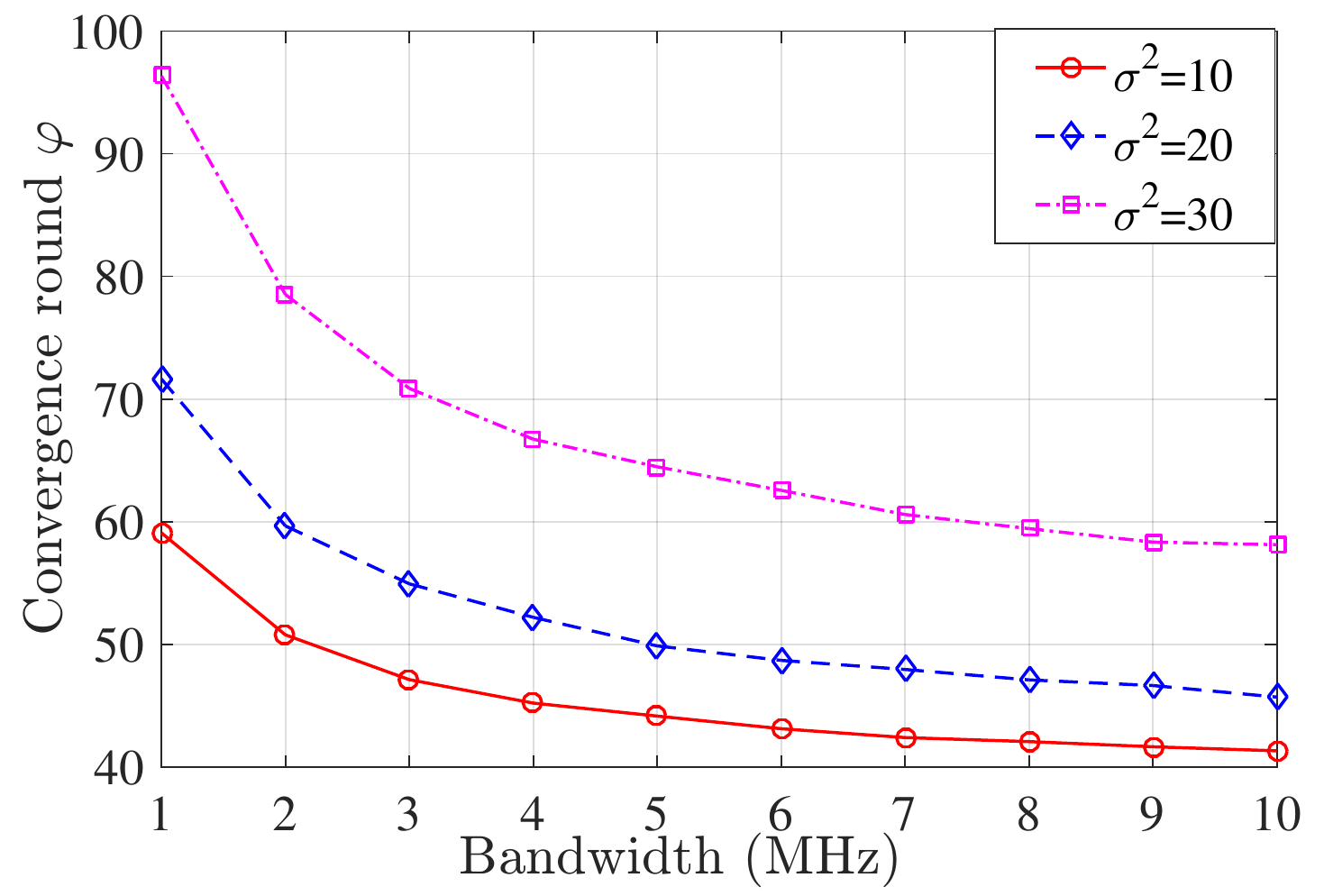}
	\DeclareGraphicsExtensions.
	\caption{\small Impact of angle deviations on the FL convergence.}
	\label{wind}
\end{figure}

Fig. \ref{wind} shows the convergence round when the variance of angle deviations changes. 
From Fig. \ref{wind}, we observe that, when the variance of angle deviations increases, FL needs more communication rounds to converge. 
This is due to the fact that, when the angle deviation variance increases, the antennas at transmitter and receiver in the network will be less aligned, leading to a drop in the antenna gains' product between transmitter and receiver in (\ref{uplink}) and (\ref{downlink}).
As a result, the transmission delay of wireless links will increase, and the probability of meeting the delay requirements, i.e., $\mathbb{P}(T_{iL,t}\!\!\leq\!\! T_{u},T_{Li,t}\!\!\leq\!\! T_{d})$, decreases.
Therefore, more communication rounds are needed to achieve the FL convergence.  
Moreover, as shown in Fig. \ref{wind}, when the bandwidth allocated to uplink and downlink increases, the FL algorithm requires fewer communication rounds to achieve convergence. 
This stems from the fact that, a large bandwidth improves the probability of meeting the delay requirements, yielding a fast FL convergence.

\begin{figure}[!t]
	\centering
	\includegraphics[width=2.6in,height=1.75in]{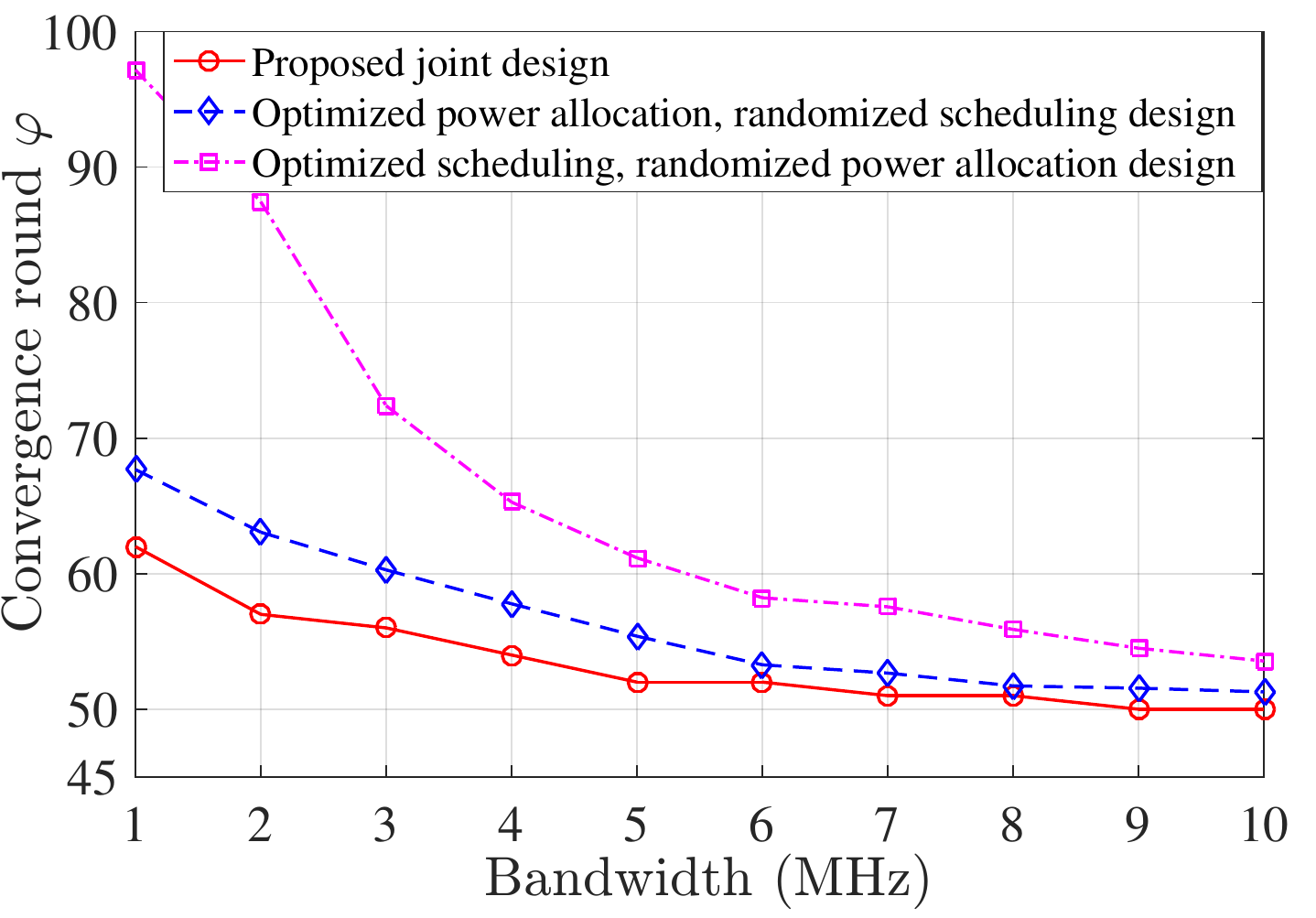}
	\DeclareGraphicsExtensions.
	\caption{\small Comparisons between systems with and without joint design.}
	\label{comparison}
\end{figure}

Fig. \ref{comparison} compares our proposed joint power allocation and scheduling design with the baselines without a joint design.
It is shown that, for the same network setting, 
the convergence round for a network with joint design is always less than its counterparts of baselines. 
In particular, when the bandwidth is $1$~MHz, the system with a joint design reduces the convergence round by as much as $35$\% compared with the baseline system with optimized scheduling and randomized power allocation design.  
Moreover, as shown in Fig. \ref{comparison}, when the bandwidth assigned to uplink and downlink increases, the performance gap between the system with the proposed joint design and the baselines decreases. 
That is because, as we increase the bandwidth, it becomes more probable for all three systems to meet the delay constraints at uplink and downlink. Therefore, the impact of communications delay on the FL convergence will be minimized.

\section{Conclusions}
In this paper, we have studied the possibility of implementing FL over a swarm of UAVs.
In particular, we have carried out a convergence analysis to study the impact of wireless factors, such as transmission delay and antenna angle deviations, on the convergence of FL. 
Using the derived insight, we have jointly designed the power allocation and scheduling policy for the UAV swarm to optimize the convergence performance of FL while guaranteeing the stability of control system and controlling the energy consumption. 
Simulation results have corroborated the convergence analysis of FL and showed the merits of the proposed joint design. 

\appendix
\subsection{Proof of Theorem \ref{ref1}}
\label{prooffortheorem6}
According to the assumptions about function $F(\boldsymbol{w}):\mathbb{R}^{n}\rightarrow \mathbb{R}$ made in Section \uppercase\expandafter{\romannumeral3}, we know that function $F(\boldsymbol{w})$ is continuously differentiable, and the gradient of $F(\boldsymbol{w})$ is uniformly Lipschitz continuous, i.e., for some positive parameter $U$, $||\triangledown F(\boldsymbol{w}^{(t+1)}) - \triangledown F(\boldsymbol{w}^{(t)})|| \leq U ||\boldsymbol{w}^{(t+1)}-\boldsymbol{w}^{(t)}||$; the function $F$ is strongly convex with positive parameter $\mu$: $F(\boldsymbol{w}^{(t+1)})\geq F(\boldsymbol{w}^{(t)})+(\boldsymbol{w}^{(t+1)}-\boldsymbol{w}^{(t)})^T \triangledown F(\boldsymbol{w}^{(t)}) + \frac{1}{2}\mu ||\boldsymbol{w}^{(t+1)}-\boldsymbol{w}^{(t)}||$.
If $F$ is twice-continuously differentiable, these two assumptions are equivalent to $\mu \boldsymbol{I} \leq \triangledown^{2}F(\boldsymbol{w}) \leq U \boldsymbol{I}$. 
Also, following a standard assumption in stochastic optimization, we consider that there exists constants $\zeta_{1} \geq 0$ and $\zeta_{2}\geq 1$, meeting $||\triangledown F_{i}(\boldsymbol{w})||^2 \leq \zeta_{1} + \zeta_{2} ||\triangledown F(\boldsymbol{w})||^2$ \cite{bertsekas1996neuro}.  

In this case, since the global FL model is the aggregation of all local FL models, the global FL model without the impact of the transmission delay can be given as 
\begin{align}
\boldsymbol{w}^{(t)} = \frac{\sum_{i=1}^{I}N_{i}\boldsymbol{w}^{(t)}_{i}}{\sum_{i=1}^{I}N_{i}}= \boldsymbol{w}^{(t-1)}-\lambda \triangledown F(\boldsymbol{w}^{(t-1)}).
\end{align}	
After taking into account the impact of transmission delays, we can rewrite the global FL model update as 
\begin{align}
\boldsymbol{w}^{(t)} = \frac{\sum_{i=1}^{I}N_{i}\boldsymbol{w}^{(t)}_{i}C_{i,t}}{\sum_{i=1}^{I}N_{i}C_{i,t}}=\boldsymbol{w}^{(t-1)}-\lambda (\triangledown F(\boldsymbol{w}^{(t-1)})+e^{(t)}), \nonumber
\end{align}
where $e^{(t)}=-\triangledown F(\boldsymbol{w}^{(t-1)})+\frac{\sum_{i=1}^{I}N_{i}\triangledown F_{i}(\boldsymbol{w}^{(t-1)})C_{i,t}}{\sum_{i=1}^{I}N_{i}C_{i,t}}$. 
Based on the assumption on the uniform Lipschitz continuity and strong convexity,  we can have the following inequalities:
\begin{align}
	\label{4}
	F(\boldsymbol{w}^{(t)}) &\leq F(\boldsymbol{w}^{(t-1)}) + (\boldsymbol{w}^{(t)}-\boldsymbol{w}^{(t-1)})^{T}\triangledown F(\boldsymbol{w}^{(t-1)})  \nonumber \\ 
	&+ \frac{U}{2} ||\boldsymbol{w}^{(t)}-\boldsymbol{w}^{(t-1)}||^2, \\ 
	\label{5}
	F(\boldsymbol{w}^{(t)}) &\geq F(\boldsymbol{w}^{(t-1)}) + (\boldsymbol{w}^{(t)}-\boldsymbol{w}^{(t-1)})^{T}\triangledown F(\boldsymbol{w}^{(t-1)}) \nonumber \\ 
	&+ \frac{\mu}{2} ||\boldsymbol{w}^{(t)}-\boldsymbol{w}^{(t-1)}||^2.
	\end{align}
	Since $\boldsymbol{w}^{(t)}=\boldsymbol{w}^{(t-1)}-\lambda(\triangledown F(\boldsymbol{w}^{(t-1)})+e^{(t)})$, we can simplify (\ref{4}) when the learning rate is $\lambda=\frac{1}{U}$ as  
	\begin{align}
	\label{6}
	F(\boldsymbol{w}^{(t)}) &\leq F(\boldsymbol{w}^{(t-1)}) \!-\! \frac{1}{U} (\triangledown F(\boldsymbol{w}^{(t-1)})+e^{(t)})^{T}\triangledown F(\boldsymbol{w}^{(t-1)}) \nonumber \\  &\hspace{0.1in}+ \frac{1}{2U} ||\triangledown F(\boldsymbol{w}^{(t-1)})+e^{(t)}||^2\nonumber \\ 
	&= F(\boldsymbol{w}^{(t-1)})-\frac{1}{2U} ||\triangledown F(\boldsymbol{w}^{(t-1)}) ||^2 + \frac{1}{2U} ||e^{(t)}||^2.
	\end{align}
	To find a lower bound on the norm of $\triangledown F(\boldsymbol{w}^{(t)})$, we can minimize both sides of (\ref{5}) with respect $\boldsymbol{w}^{(t)}$. 
The minimal value of the left-hand side of (\ref{5}) is achieved when $\boldsymbol{w}^{(t)} = \boldsymbol{w}^{*}$, and the minimal value of the right-hand side of (\ref{5}) is realized when $\boldsymbol{w}^{(t)} = \boldsymbol{w}^{(t-1)} - \frac{1}{\mu} \triangledown F(\boldsymbol{w}^{(t-1)})$. 
Particularly, we have  
\begin{align}
\label{7}
F(\boldsymbol{w}^{*}) \geq F(\boldsymbol{w}^{(t-1)}) - \frac{1}{2 \mu} ||\triangledown F(\boldsymbol{w}^{(t-1)})||^2.
\end{align}
When replacing $\boldsymbol{w}^{(t-1)}$ with $\boldsymbol{w}^{(t)}$ in (\ref{7}), we can obtain a lower bound for the norm of $\triangledown F(\boldsymbol{w}^{(t)})$ as 
\begin{align}
\label{8}
||\triangledown F(\boldsymbol{w}^{(t)})||^2 \geq 2 \mu (F(\boldsymbol{w}^{(t)})-F(\boldsymbol{w}^{*})).
\end{align}
Combining (\ref{6}) and (\ref{8}), we can obtain an upper bound of the current loss and the minimal loss given by 
\vspace{-7pt}
\par \nobreak
{\small\begin{align}
	F(\boldsymbol{w}^{(t)}) - F(\boldsymbol{w}^{*}) \leq (1- \frac{\mu}{U})[F(\boldsymbol{w}^{(t-1)})-F(\boldsymbol{w}^{*})] + \frac{1}{2U} ||e^{(t)}||^2. \nonumber
	\end{align}}According to \cite{friedlander2012hybrid}, when $\mathbb{E}[||e^{(t)}||^2] \leq 2U(\frac{\mu}{U}-\rho^{(t)})\mathbb{E}(F(\boldsymbol{w}^{(t)})-F(\boldsymbol{w}^{*}))$, we can achieve the strong expected linear convergence, i.e., 
\begin{align}
\mathbb{E}(F(\boldsymbol{w}^{(t)}) - F(\boldsymbol{w}^{*})) \leq (1-\rho^{(t)}) \mathbb{E}(F(\boldsymbol{w}^{(t-1)}) - F(\boldsymbol{w}^{*})). \nonumber 
\end{align}
According to the strong expected linear convergence requirement, we know that the convergence rate satisfies 
\begin{align}
\label{convergence_inequality}
\rho^{(t)} \leq \frac{\mu}{U} - \frac{\mathbb{E}[||e^{(t)}||^2] }{2U \mathbb{E}(F(\boldsymbol{w}^{(t)})-F(\boldsymbol{w}^{*}))}.
\end{align} 
By using the results in \cite{chen2019joint}, we have the following inequality: 
\begin{align}
\label{32}
\mathbb{E}(||e^{(t)}||^2) \leq &\frac{1}{N}\sum_{i=1}^{I}N_{i}(\zeta_{1}+\zeta_{2}\mathbb{E}(\triangledown F(\boldsymbol{w}^{(t)}))) \times \nonumber \\ &(1\!-\!\mathbb{P}(T_{iL}\!\leq\! T_{u}(\beta),T_{Li}\!\leq\! T_{d}(\beta))).
\end{align}
The right-hand side of (\ref{convergence_inequality}) will meet the following inequality: 
\begin{align}
\label{convergence_inequality1}
&\frac{\mu}{U} \!\!-\!\! \frac{\mathbb{E}[||e^{(t)}||^2] }{2U \mathbb{E}(F(\boldsymbol{w}^{(t)})\!\!-\!\!F(\boldsymbol{w}^{*}))} \geq \frac{\mu}{U}\!\!-\!\!\sum_{i=1}^{I}\!\!N_{i}(\zeta_{1}\!\!+\!\!\zeta_{2}\mathbb{E}(\triangledown\! F(\boldsymbol{w}^{(t)})))  \nonumber \\ 
&\hspace{0.5in}\times \frac{  (1\!-\!\mathbb{P}(T_{iL}\!\leq\! T_{u}(\beta),T_{Li}\!\leq\! T_{d}(\beta)))  }{2NU \mathbb{E}(F(\boldsymbol{w}^{(t)})\!-\!F(\boldsymbol{w}^{*}))}.
\end{align} 
Therefore, to guarantee that (\ref{convergence_inequality}) always exists, we have 
	\begin{align}
	&\rho^{(t)} \leq \frac{\mu}{U}-\sum_{i=1}^{I}\!N_{i}(\zeta_{1}\!+\!\zeta_{2}\mathbb{E}(\triangledown\! F(\boldsymbol{w}^{(t)})))  \nonumber \\ 
	&\hspace{0.5in}\times \frac{  (1\!-\!\mathbb{P}(T_{iL}\!\leq\! T_{u}(\beta),T_{Li}\!\leq\! T_{d}(\beta)))  }{2NU \mathbb{E}(F(\boldsymbol{w}^{(t)})\!-\!F(\boldsymbol{w}^{*}))}\nonumber \\ 
	&\stackrel{(a)}{\leq} \!\frac{\mu}{U}-\sum_{i=1}^{I}\!N_{i}(\zeta_{1}\!+\!\zeta_{2}2 \mu (F(\boldsymbol{w}^{(t)})\!-\!F(\boldsymbol{w}^{*})))\nonumber \\ 
	&\hspace{0.5in}\times \!\frac{  (1\!\!-\!\!\mathbb{P}(T_{iL}\!\leq\! T_{u}(\beta),T_{Li}\!\leq\! T_{d}(\beta)))    }{2NU \mathbb{E}(F(\boldsymbol{w}^{(t)})\!-\!F(\boldsymbol{w}^{*}))}\nonumber \\ 
	&\stackrel{(b)}{\leq} \frac{\mu}{U}-\sum_{i=1}^{I}N_{i}(2 \mu \mathbb{E}(F(\boldsymbol{w}^{(t)})-F(\boldsymbol{w}^{*}))) \nonumber \\
	&\hspace{0.5in}\times\frac{  (1\!-\!\mathbb{P}(T_{iL}\!\leq\! T_{u}(\beta),T_{Li}\!\leq\! T_{d}(\beta)))    }{2NU \mathbb{E}(F(\boldsymbol{w}^{(t)})-F(\boldsymbol{w}^{*}))}\nonumber \\
	& = \frac{\mu}{U}-\frac{ \sum_{i=1}^{I}N_{i} \mu (1\!-\!\mathbb{P}(T_{iL}\!\leq\! T_{u}(\beta),T_{Li}\!\leq\! T_{d}(\beta)))}{NU}, 
	\end{align}
	where in (a), we use the results derived in (\ref{8}), and the derivation in (b) is based on the fact that $\zeta_{1}\geq 0$ and $\zeta_{2}\geq 1$.
Assume $\rho = \frac{\mu}{U}-\frac{ \sum_{i=1}^{I}N_{i} \mu (1\!-\!\mathbb{P}(T_{iL}\!\leq\! T_{u}(\beta),T_{Li}\!\leq\! T_{d}(\beta)))}{NU}=\frac{ \sum_{i=1}^{I}N_{i} \mu (\mathbb{P}(T_{iL}\!\leq\! T_{u}(\beta),T_{Li}\!\leq\! T_{d}(\beta)))}{NU}$, then, we can have 
\begin{align}
\mathbb{E}(F(\boldsymbol{w}^{(t)}) - F(\boldsymbol{w}^{*})) &\leq (1-\rho) \mathbb{E}(F(\boldsymbol{w}^{(t-1)}) - F(\boldsymbol{w}^{*})) \nonumber \\ 
& \leq (1-\rho)^2 \mathbb{E}(F(\boldsymbol{w}^{(t-2)}) - F(\boldsymbol{w}^{*}))\nonumber \\ 
& ... \nonumber \\ 
&  \leq (1-\rho)^t \mathbb{E}(F(\boldsymbol{w}^{(0)}) - F(\boldsymbol{w}^{*})). \nonumber 
\end{align}
We can further determine the convergence round needed to achieve a target difference threshold, i.e., $\mathbb{E} (F(\boldsymbol{w})-F(\boldsymbol{w}^{*}))\leq \varepsilon$, as follows:
\begin{align}
t &\geq \log_{1-\rho} \frac{\varepsilon}{\mathbb{E}(F(\boldsymbol{w}^{(0)}) - F(\boldsymbol{w}^{*}))} \nonumber \\
&\stackrel{(a)}{\geq}\log_{1-\rho} \frac{\varepsilon}{\mathbb{E}(F(\boldsymbol{w}^{(0)}))} \nonumber \\ 
&=\log_{1-\rho} \frac{\varepsilon}{\sum_{i=1}^{I}\sum_{n=1}^{N_{i}}f(\boldsymbol{w}^{(0)},\boldsymbol{x}_{in},y_{in})},
\end{align} 
where in (a), we use the fact that $1-\rho \leq 1$.
Since the convergence round must be integral, we can have the results in Theorem 1.

	\def\baselinestretch{1}
\bibliographystyle{IEEEtran}

\end{document}